\documentclass[a4paper,conference]{IEEEtran}
\usepackage{cite}
\usepackage{amsmath,amssymb,amsfonts}
\usepackage{graphicx}
\usepackage{subfigure}
\usepackage{textcomp}
\usepackage{xcolor}
\usepackage{makecell}
\usepackage{CJKutf8}\usepackage{algorithm}  
\usepackage{algpseudocode}  
\usepackage{amsmath}
\usepackage{shadowtext}
\usepackage{color, colortbl}
\usepackage{multirow}
\definecolor{Gray}{gray}{0.9}

\ifCLASSINFOpdf
\else
\fi
\usepackage{url}

\begin{document}
%
\title{GFTE: Graph-based Financial Table Extraction}



%
\author{\IEEEauthorblockN{Yiren Li\IEEEauthorrefmark{1},
Zheng Huang\IEEEauthorrefmark{2},
Junchi Yan\IEEEauthorrefmark{3},
Yi Zhou\IEEEauthorrefmark{4},
Fan Ye\IEEEauthorrefmark{5} and
Xianhui Liu\IEEEauthorrefmark{6}}
\IEEEauthorblockA{\IEEEauthorrefmark{1}\IEEEauthorrefmark{2}\IEEEauthorrefmark{3}\IEEEauthorrefmark{4}School of Electronic Information and Electrical Engineering\\
Shanghai Jiao Tong University,
Email: \{irene716, huang-zheng, yanjunchi, zy\_21th\}@sjtu.edu.cn}
\IEEEauthorblockA{\IEEEauthorrefmark{5}\IEEEauthorrefmark{6}China Financial Fraud Research Center\\
Email: yefan@xnai.edu.cn, lxianhui@tjzxcn.com}}


\maketitle

\begin{abstract}
Tabular data is a crucial form of information expression, which can organize data in a standard structure for easy information retrieval and comparison. However, in financial industry and many other fields tables are often disclosed in unstructured digital files, e.g. Portable Document Format (PDF) and images, which are difficult to be extracted directly. In this paper, to facilitate deep learning based table extraction from unstructured digital files, we publish a standard Chinese dataset named FinTab, which contains more than 1,600 financial tables of diverse kinds and their corresponding structure representation in JSON. In addition, we propose a novel graph-based convolutional neural network model named GFTE as a baseline for future comparison. GFTE integrates image feature, position feature and textual feature together for precise edge prediction and reaches overall good results \footnote{https://github.com/Irene323/GFTE}.
\end{abstract}


%
\IEEEpeerreviewmaketitle

\renewcommand{\arraystretch}{1.5}
\section{Introduction}
In the information age, how to quickly obtain information and extract key information from massive and complex resources has become an important issue \cite{B40}. In the meantime, with the increase in the number of enterprises and the growing amount of disclosure of financial information, extracting key information has also become an essential means to improve the efficiency in financial information exchange process \cite{B41, B42}. In recent years, some new researches have also begun to focus on improving the efficiency and accuracy of information retrieval technology \cite{B11,B12}.

Table, as a form of structured data, is both simple and standardized. Hurst et al. \cite{B13} regard a table as a representation of a set of relations between organized hierarchical concepts or categories, while Long et al. \cite{B14} consider it as a superstructure imposed on a character-level grid. Due to its clear structure, table data can be quickly understood by users. Financial data, especially digital information, are often presented in tabular form. In a manner of speaking, table data, as key information in financial data, are increasingly valued by financial workers during financial data processing.

Although table extraction is a common task in various domains, extracting tabular information manually is often a tedious and time-consuming process. We thus require automatic table extraction methods to avoid manual involvement. However, it is still difficult for the existing methods to accurately recover the structure of relatively complicated financial tables.

\begin{figure}[htbp]
\centering
\subfigure[The ground truth structure of part of a financial table. Each cell in the first two lines is marked with different colors.]{
\begin{minipage}[t]{\linewidth}
\centering
\includegraphics[width=\linewidth]{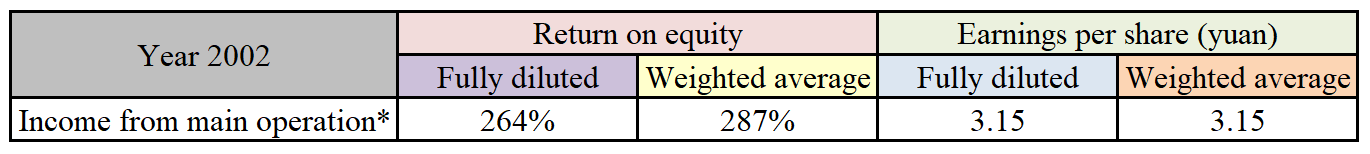}
\end{minipage}%
}%

\subfigure[The recognized structure by Adobe Acrobat DC.]{
\begin{minipage}[t]{\linewidth}
\centering
\includegraphics[width=\linewidth]{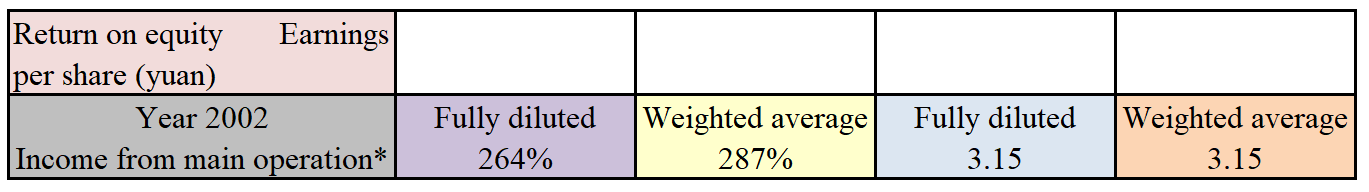}
\end{minipage}%
}%

\subfigure[The recognized structure using Tabby \cite{B3}.]{
\begin{minipage}[t]{\linewidth}
\centering
\includegraphics[width=\linewidth]{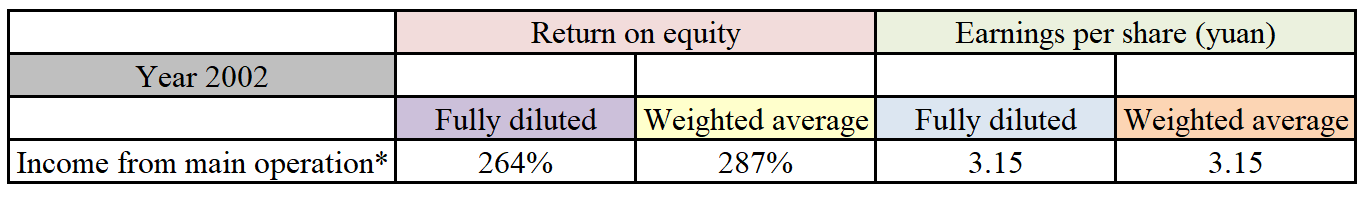}
\end{minipage}
}%
\centering
\caption{An example of a table with spanned cells and the recovered table structures with the existing methods.}
\label{fig::existing}
\end{figure}

Figure \ref{fig::existing} illustrates an intuitive example of the performance of different existing methods, i.e. Adobe Acrobat DC and Tabby \cite{B3}. Both of them fail to give the correct result. Meanwhile, it is not hard to notice that problems often occur at spanning cells, which very likely carry the information of table headers and are thus critical for table extraction and understanding. Therefore, the performance of table extraction methods are still hoped to be improved, especially by the complicated cases.

Based on these considerations, since the design of artificial intelligence algorithms relies on standard data and test benchmarks, we construct an open source financial benchmark dataset named FinTab. More specifically, sample collection, sample sorting and cleaning, benchmark data determining and baseline method test were completed. FinTAb can be further used in financial context in terms of table extraction, key information extraction, image data identification, bill identification and other specific content. With a more comprehensive benchmark dataset, we hope to promote the emergence of more innovative technologies. Further detailed information about our standard financial dataset will be introduced in Section \ref{sec::dataset}.

Besides, this paper also proposes a novel table extraction method, named GFTE, with the help of graph convolutional network (GCN). GFTE can be used as a baseline, which regards the task of table structure recognition as an edge prediction problem based on graph. More specifically, we integrate image feature, textual feature and position feature together and feed them to a GCN to predict the relation between two nodes. Details about this baseline algorithm will be discussed in section \ref{sec::algo}.

In general, the contributions of this work can be summarized as following:
\begin{enumerate}
    \item A Chinese benchmark dataset FinTab of more than 1,600 tables of various difficulties, containing table location, structure identification and table interpretation information.
    \item We propose a graph-based convolutional network model named GFTE as table extraction baseline. Extensive experiments demonstrate that our proposed model outperforms state-of-the-art baselines greatly.
\end{enumerate}

\section{Related Work}\label{sec::related_work}

In this section, we will first familiarize the reader with some previous published datasets and some related contests, and then present a overview of table extraction technologies.

\subsection{Previous datasets}

We introduce some existing public available datasets:

\textbf{Marmot} The Marmot dataset \cite{B34} is composed of both Chinese and English pages. The Chinese pages are collected from over 120 e-Books in diverse fields of subject provided by Founder Apabi library, while the English pages are from Citeseer website. Derived from PDF, the dataset stores tree structure of all document layouts, where the leaves are characters, images and paths, and the root is the whole page. Internal nodes include textlines, paragraphs, tables etc.

\textbf{UW3 and UNLV} The UW3 dataset \cite{B35} is collected from 1,600 pages of skew-corrected English document and 120 of them contain at least one marked table zone. THe UNLV dataset derives from 2,889 pages of scanned document images, in which 427 images include table.

\textbf{ICDAR 2013} This dataset \cite{B30} includes a total of 150 tables: 75 tables in 27 excerpts from the EU and 75 tables in 40 excerpts from the US Government, i.e. in total 67 PDF documents with 238 pages in English.

\textbf{ICDAR 2019} This dataset for the ICDAR 2019 Competition on Table Detection and Recognition \cite{B36} is separated into training part and test part. The training dataset contains images of 600 modern documents and their bounding boxes of table region, as well as images of 600 archival documents, their table structures and bounding boxes of both table region and cell region. In the test dataset, images and table regions of 199 archival documents and 240 modern ones are offered. Besides, table structures and cell regions of 350 archival documents are also included.

\textbf{PubTabNet} The PubTabNet dataset \cite{B37} contains more than 568 thousand images of tabular data annotated with the corresponding HTML representation of the tables. More specifically, table structure and characters are offered but the bounding boxes are missing.

\textbf{SciTSR} SciTSR \cite{B38} is a comprehensive dataset, which consists of 15,000 tables in PDF format, images of the table region, their corresponding structure labels and bounding boxes of each cell. It is split into 1,2000 for training and 3,000 for test. Meanwhile, a list of complicated tables, called SciTSR-COMP, is also provided.

\textbf{TableBank} The TableBank \cite{B39} is image-based table detection and recognition dataset. Since two tasks are involved, it is composed of two parts. For the table detection task, images of the pages and bounding boxes of tables region are included. For the table structure recognition task, images of the page and HTML tag sequence that represents the arrangement of rows and columns as well as the type of table cells are provided. However, textual content recognition is not the focus of this work, so textual content and its bounding boxes are not contained.

Table \ref{tab::previous_datasets} gives more information for comparison.

\begin{table*}[htbp]
\caption{Public datasets for table recognition}
\begin{center}
\begin{tabular}{|l|p{4cm}|p{7cm}|p{1cm}|l|p{1cm}|}
\hline
Name & Source & Content & Amount & Format & Language\\
\hline
Marmot & e-Books and Citeseer website & tree structure of all document layouts & 2,000 & bmp, xml & Chinese, English\\
\hline
UW3 & skew-corrected document & images of page, manually edited bounding boxes of page frame, text and non-text zones, textlines, and words, type of each zone (text, math, table, half-tone, ...)  & 120 & png, xml & English \\
\hline
UNLV & magazines, news papers, business letter, annual Report, etc. & scanned images of page, bounding boxes for rows, columns and cells, OCR recognized words within the whole page & 427 & png, xml & English \\
\hline
ICDAR 2013 & European Union and US Government websites & PDF documents, bounding boxes of table, textual content, structure labels, bounding boxes for each cell & 150 & pdf, xml & English \\
\hline
ICDAR 2019 & modern documents and archival ones with various formats & 840 jpgs and xmls including bounding boxes of table in modern documents, 1149 jpgs and xmls including structure labels, bounding boxes of table and bounding boxes for each cell in archival documents & about 2,000 & jpg, xml & English \\
\hline
PubTabNet & scientific publications included in the PubMed Central Open Access Subset & images of tabular data, textual content, table structure labels in HTML & more than 568,000 & png, json & English \\
\hline
SciTSR & LaTeX source files & PDF documents, images, textual content, structure labels, bounding boxes for each cell & 15,000 & png, json & English \\
\hline
TableBank & Word and Latex documents on the internet such as business documents, official fillings, research papers, etc. & Table Detection: bounding boxes of table; Table Structure Recognition: table structure labels & 417,234; 145,463 & jpg, HTML & English \\ 
\hline
FinTab & Annual and semi-annual reports, debt financing, bond financing, collection of medium-term notes, short-term financing, prospectus & textual ground truth, structure information and the unit of the table & 1,685 & pdf, json & Chinese \\
\hline
\end{tabular}
\label{tab::previous_datasets}
\end{center}
\end{table*}

\subsection{Methods}

Table extraction is considered as a part of table understanding \cite{B2}, and conventionally consists of two steps \cite{B3}: 
\begin{enumerate}
    \item Table Detection. Namely, a certain part of the file is identified as a table in this step.
    \item Table Structure Decomposition. This task aims to recover the table into its components as close to the original as possible. For example, the proper identification of header elements, the structure of columns and rows, correct allocation of data units, etc.
\end{enumerate}

In the past two decades, a few methods and tools have been devised for table extraction. Some of them are discussed and compared in some recent surveys \cite{B7, B4, B8, B9, B5, B10}.

There are generally three main categories of the existing approaches \cite{B15}:
\begin{enumerate}
    \item Predefined layout-based approaches,
    \item Heuristic-based approaches,
    \item Statistical or optimization-based approaches.
    
\end{enumerate}

\textbf{Predefined layout-based approaches} design several templates for possible table structures. Certain parts of the document is identified as tables, if they correspond to certain templates. Shamilian \cite{B16} proposes a predefined layout-based table identification and segmentation algorithm as well as a graphical user interface (GUI) for defining new layouts. Nevertheless, it only works well in single-column cases. A wrapper-based approach, which transforms the low-level PDF instructions into text segments, is mentioned in \cite{B17}. Mohemad et al. \cite{B18} present another predefined layout-based approach, which focuses on paragraph and tabular, then associated text using a combination of heuristic, rule-based and predefined indicators. However, a disadvantage of these approaches is that tables can only be classified into the previous defined layouts, while there are always limited types of templates defined in advance.

\textbf{Heuristic-based approaches} specify a set of rules to make decisions so as to detect tables which meet certain criteria. According to \cite{B9}, heuristics-based approaches remain dominant in literature. \cite{B19} is the first relative research focusing on PDF table extraction, which uses a tool named pdf2hmtl to return text pieces and their absolute coordinates, and then utilizes them for table detection and decomposition. This technique achieves good results for lucid tables, but it is limited as it assumes all pages to be single column. Liu et al. \cite{B20} propose a set of medium-independent table metadata to facilitate the table indexing, searching, and exchanging, in order to extract the contents of tables and their metadata. Another heuristic-based approach \cite{B21} present a bottom-up way to identify tabular arrangements. Basic content elements are grouped up according to their spatial features.

\textbf{Statistical approaches} make use of statistical measures obtained through offline training. The estimated parameters are then taken for practical table extraction. Different statistical models have been used, for example, probabilistic modelling \cite{B22}, the Naive Bayes classifier \cite{B23, B24}, decision trees \cite{B25, B26}, Support Vector Machine \cite{B25, B27}, Conditional Random Fields \cite{B27, B28, B29}, graph neural network \cite{B38, B43, B44}, attention module \cite{B45}, etc. \cite{B46} uses a pair of deep learning models (Split and Merge models) to recover tables from images.

\begin{figure*}[htbp]
\centerline{\includegraphics[width=0.9\linewidth]{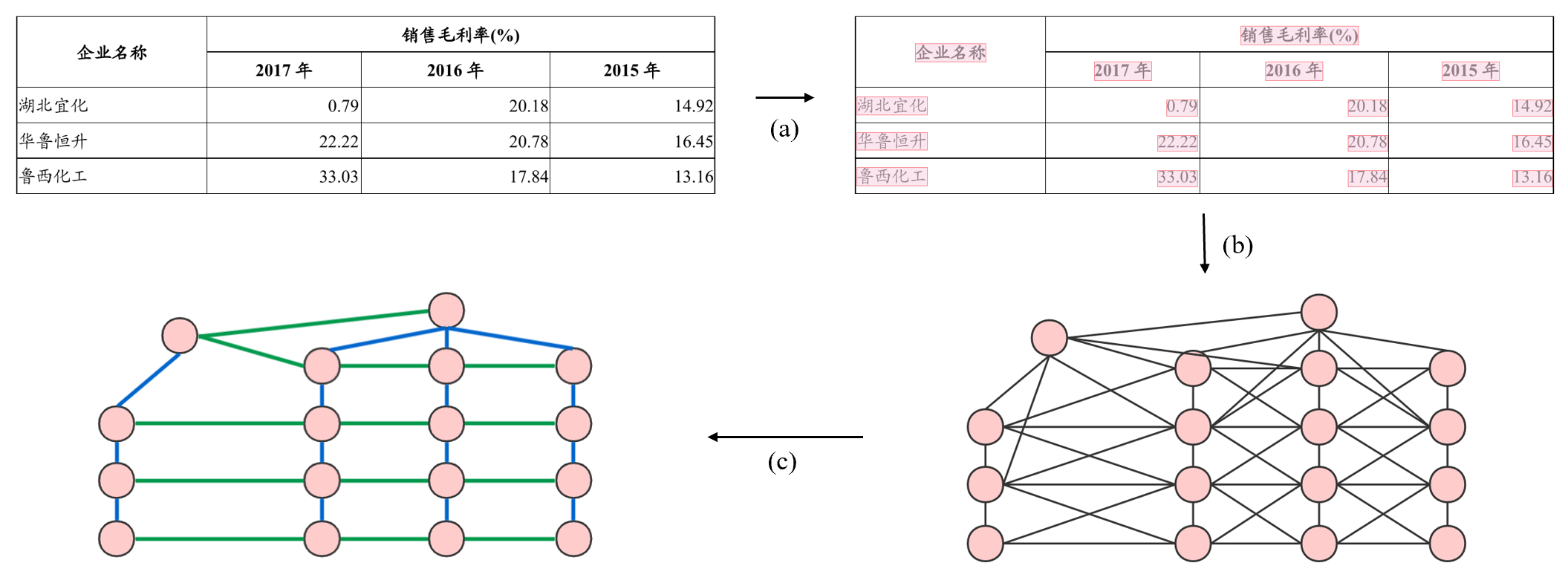}}
\caption{Overview of our novel GCN-based algorithm.}
\label{fig::overview}
\end{figure*}

\section{Dataset collection and annotation}\label{sec::dataset}

In general, there are currently following problems with the existing contests and standard datasets:
\begin{enumerate}
    \item There are few competitions and standard datasets for extracting table information from financial documents.
    \item The source for tabular information extraction lacks diversity.
\end{enumerate}

In consideration of this, the benchmark dataset FinTab released this time aims to make certain contribution in this field. In this dataset, we collect a total of 19 PDF files with more than 1,600 tables. The specific document classification is shown in Table \ref{tab::docclass}. All documents add up to 3,329 pages, while 2,522 of them contain tables. 

\begin{table}[htbp]
\caption{document classification of our benchmark dataset FinTab}
\begin{center}
\begin{tabular}{|p{4cm}|p{2cm}<{\centering}|}
\hline
File Type & Number of files \\
\hline
Annual and semi-annual reports & 3 \\
\hline
Debt financing & 2\\
\hline
Bond financing & 3\\
\hline
Collection of medium-term notes & 2\\
\hline
Short-term financing & 8\\
\hline
Prospectus & 1\\
\hline
\end{tabular}
\label{tab::docclass}
\end{center}
\end{table}

To ensure that the types of forms are diverse, in addition to the basic forms of table, special cases with different difficulties are also included, e.g. semi-ruled table, cross-page table, table with merged cells, multi-line header table, etc. It is also worth mentioning that there are 119,021 cells in total, while the number of merged cells is 2,859, accounting for 2.4\%. Detailed types and quantity distribution of tables are shown in Table \ref{tab::tabletype}.

\begin{table}[htbp]
\caption{document classification of our benchmark dataset FinTab}
\begin{center}
\begin{tabular}{|p{3.7cm}|p{2cm}<{\centering}|p{1.6cm}<{\centering}|}
\hline
\textbf{Table Type} & \textbf{Number of tables} & \textbf{Percentage} \\
\hline
Single-page table & 523 & 62.5\% \\
\hline
Double-page table & 523 & 31\% \\
\hline
Multi-page table & 108 & 6.5\% \\
\hline
Table with incomplete form line & 140-150 & 8.3\%-8.9\% \\
\hline
Table with merged cells & 583 & 34.59\% \\
\hline
Total & 1,685 & 100\% \\
\hline
\end{tabular}
\label{tab::tabletype}
\end{center}
\end{table}

FinTab contains various types of tables. Here, we briefly introduce some of them in order of difficulty.
\begin{enumerate}
    \item Basic single-page table. This is the most basic type of table, which takes up less than one page and does not include merged cells. It is worth mentioning that we offer not only textual ground truth and structure information, but also the unit of the table, because mostly financial table contains quite a few numbers.
    \item Table with merged cells. In this case, the corresponding merged cells should be recovered.
    \item Cross-page table. If the table appears to spread across pages, the cross-page table need to be merged into a single form. If the header of the two pages appears to be duplicated, only one needs to be remained. Page number and other useless information should also be removed. Another difficult situation to be noticed is that if a single cell is separated by two pages, it should be merged into one according to its semantics.
    \item Table with incomplete form line. In this case, it is necessary to intelligently locate the dividing line according to the position, format, and meaning of the text.
\end{enumerate}


\begin{figure*}[htbp]
\centerline{\includegraphics[width=0.9\linewidth]{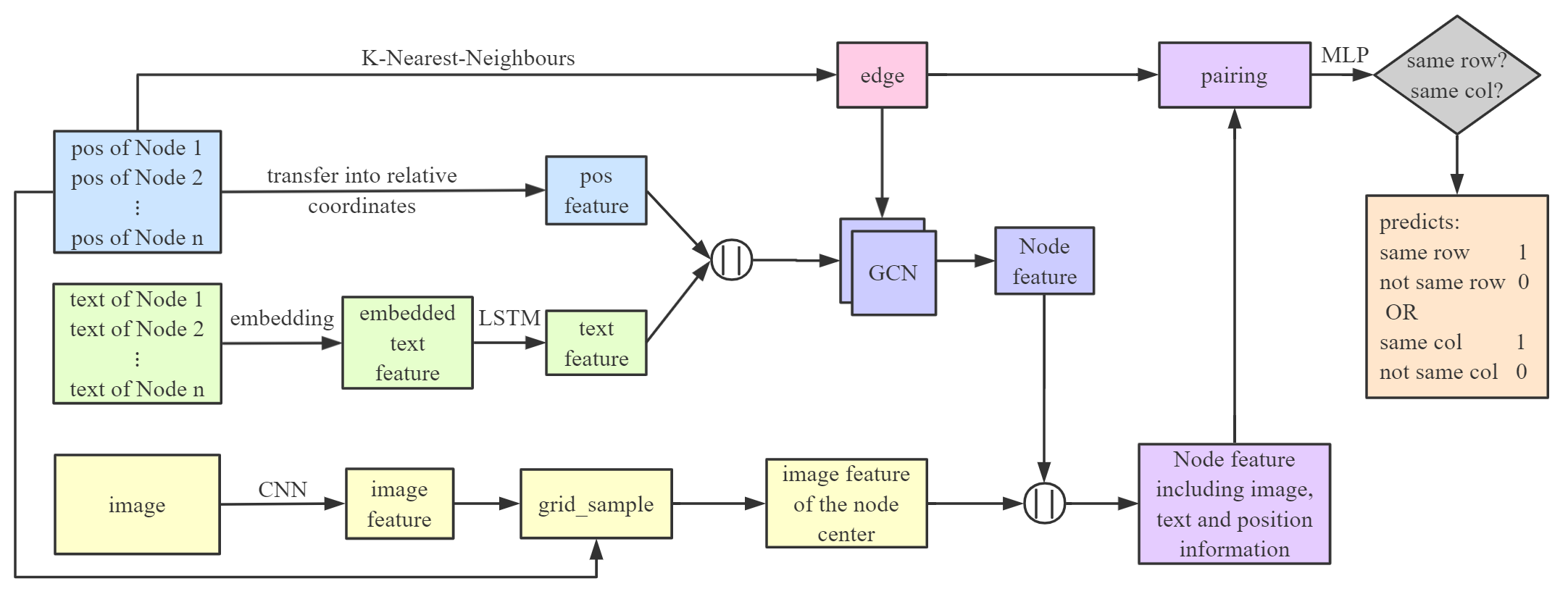}}
\caption{The structure of our proposed GCN-based algorithm GFTE.}
\label{fig::net}
\end{figure*}

\section{Baseline algorithm}\label{sec::algo}

In this paper, we also propose a novel graph-neural-network-based algorithm named GFTE to fulfill table structure recognition task, which can be used as a baseline. In this section, we introduce detailed procedure of this algorithm.

Figure \ref{fig::overview} illustrates an overview of GFTE. Since our dataset is in Chinese, we give a translated version of the example in Table \ref{tab::translation} for better understanding. Given a table in PDF document, our method can be summarized into the following steps:
\renewcommand{\theenumi}{(\alph{enumi}}
\begin{enumerate}
    \item We build its ground truth, which consists of (1) image of the table region, (2) textual content, (3) text position and (4) structure labels. \label{item:first}
    \item Then, we construct an undirected graph $G=<V, R_C>$ on the cells. \label{item:second}
    \item Finally, we use our GCN-based algorithm to predict adjacent relations, including both vertical and horizontal relations. 
\end{enumerate}

\begin{table}[htbp]
\caption{The translated version of the table we used for illustration in this paper.}
\begin{center}
\begin{tabular}{|p{2.4cm}|c|c|c|}
\hline
\multirow{2}{2.4cm}{Company Name} & \multicolumn{3}{c|}{Gross Profit Margin(\%)} \\
\cline{2-4}
~ & Year 2017 & Year 2016 & Year 2015 \\
\hline
Hubei Yihua & 0.79 & 20.18 & 14.92 \\
\hline
Hualu Hengsheng & 22.22 & 20.78 & 16.45 \\
\hline
Luxi Chemical & 33.03 & 17.84 & 13.16 \\
\hline
\end{tabular}
\label{tab::translation}
\end{center}
\end{table}

In sub-section \ref{sec::prob}, we first introduce how we comprehend this table structure recognition problem. 

\subsubsection{Problem Interpretation}\label{sec::prob}
In a table recognition problem, it is quite natural to consider each cell in the table as a node. Then, the vertical or horizontal relation between a node and its neighbors can be understood as the feature of edges.

If we use $N$ to denote the set of nodes and $E_C$ to denote the fully connected edges, then a table structure can be represented by a complete graph $G=<V, R_C>$, where $R_C$ indicate a set of relations between $E_C$. More specifically, we have $R_C = E_C \times \{vertical, horizontal, unrelated\}$.

Thus, we can interpret the problem as the following: given a set of nodes $N$ and their feature, our aim is to predict the relations $R_C$ between pairs of nodes as accurate as possible.

However, training on complete graphs is extremely expensive. It is not only computationally intensive but also quite time-consuming. Meanwhile, it is not hard to notice that a table structure can be represented by far fewer edges, as long as a node is connected to its nearest neighbors including both vertical ones and horizontal ones. With the knowledge of node position, we are also capable of recovering the table structure from these relations.

Therefore in this paper, instead of training on the complete graph with $R_C$, which is of $O(\vert N \vert^2)$ complexity, we make use of the K-Nearest-Neighbors (KNN) method to construct $R$, which contains the relations between each node and its K nearest neighbors. With the help of KNN, we can reduce the complexity to $O(K*\vert N \vert)$.

\subsubsection{GFTE}
\begin{figure}[htbp]
\centerline{\includegraphics[width=\linewidth]{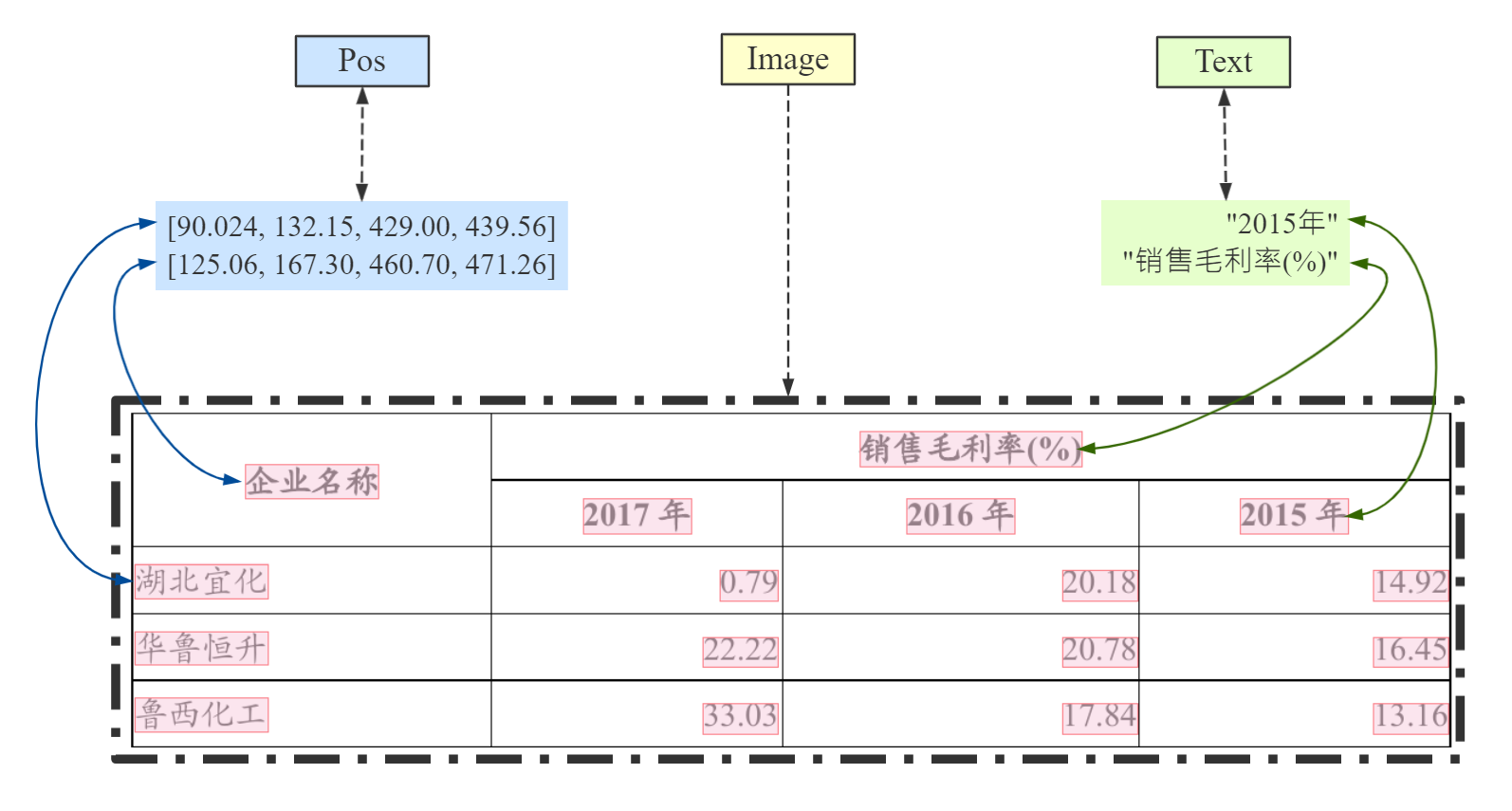}}
\caption{An intuitive example of source data format.}
\label{fig::data}
\end{figure}

\begin{figure}[htbp]
\centerline{\includegraphics[width=0.7\linewidth]{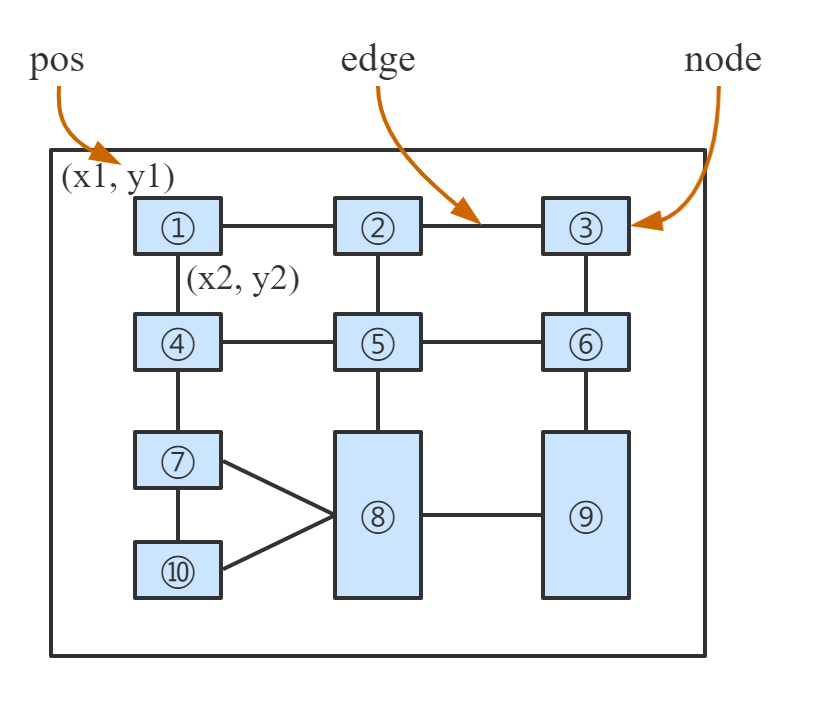}}
\caption{Ground truth structure.}
\label{fig::node}
\end{figure}

For each node, three types of information are included, i.e. the textual content, the absolute locations and the image, as shown in Figure \ref{fig::data}. We then make use of the structure relations to build the ground truth and the entire structure could be like Figure \ref{fig::node}. For higher accuracy, we train horizontal and vertical relations separately. For horizontal relations, we label each edge as \textit{$1$: in the same row} or \textit{$0$: not in the same row}. Similarly for vertical relations, we label each edge as \textit{$1$: in the same column} or \textit{$0$: not in the same column}.

Figure \ref{fig::net} gives the structure of our graph-based convolutional network GFTE. We first convert the absolute position into relative positions, which are further used to generate the graph. In the mean time, plain text is first embedded into a predefined feature space, then LSTM is used to obtain semantic feature. We concatenate the position feature and the text feature together and feed them to a two-layer graph convolutional network (GCN).

Meanwhile, we first dilate the image by a small kernel to make the table lines thicker. We also resize the image to $256\times 256$ pixels in order to normalize the input. We then use a three-layer CNN to calculate the image feature. After that, using the relative position of the node, we can calculate a flow-field grid. By computing the output using input pixel locations from the grid, we can acquire the image feature of a certain node at a certain point.

When these three different kinds of features are prepared, we pair two nodes on an edge of the generated graph. Namely we find two nodes of one edge and concatnate their three different kinds of features together. Finally, we use MLP to predict whether the two nodes are in same row or in same column.

In this baseline algorithm, we train our network with the SciTSR training dataset, which is introduced in section \ref{sec::related_work}. The SciTSR dataset still contains a certain number of faulty samples. We use some filters to reject them.

\section{Evaluation Results}

In this section, we evaluate GFTE with prediction accuracy for both vertical and horizontal relations. Our novel FinTab dataset is used to test the performance of different GFTE model structures and the SciTSR dataset is used for validation.

\begin{table}[htbp]
\caption{Accuracy results of different GFTE models on vertical and horizontal directions.}
\begin{center}
\begin{tabular}{|p{2.8cm}|p{1.6cm}<{\centering}|p{1.6cm}<{\centering}|}
\hline
\multicolumn{1}{|m{2.8cm}}{Network} &
\multicolumn{1}{|m{1.6cm}}{Horizontal prediction} & \multicolumn{1}{|m{1.6cm}|}{Vertical\quad prediction} \\
\hline
\rowcolor{Gray}
GFTE-pos & 0.759836 & 0.842450 \\
GFTE-pos+text & 0.858675 & 0.903230 \\
\rowcolor{Gray}
GFTE & 0.861019 & 0.903031 \\
\hline
\end{tabular}
\label{tab::GFTE}
\end{center}
\end{table}

Firstly, we train GFTE-pos. Namely we use the relative position and KNN algorithm to generate graph, and we train GFTE only with the position feature. Secondly, we train the network with the position feature as well as the text feature acquired by LSTM. This model is named GFTE-pos+text. Finally, our proposed GFTE is trained by further including the image feature with the help of grid sampling.

In Table \ref{tab::GFTE}, we give the performance of different models on FinTab dataset. As listed, the accuracy shows an overall upward trend when we concatenate more kinds of features. It improves distinctly when we include text feature, namely a rise of 10\% by horizontal prediction and 5\% by vertical prediction. Further including image feature seems to help improve the performance a little, but not too much.

\begin{figure}[htbp]
\centering
\subfigure[In this table, the last column is right-aligned. However, the bottom part gives all the partners' name and their capital subscriptions. The first 8 rows are thus far apart.]{
\begin{minipage}[t]{\linewidth}
\centering
\includegraphics[width=\linewidth]{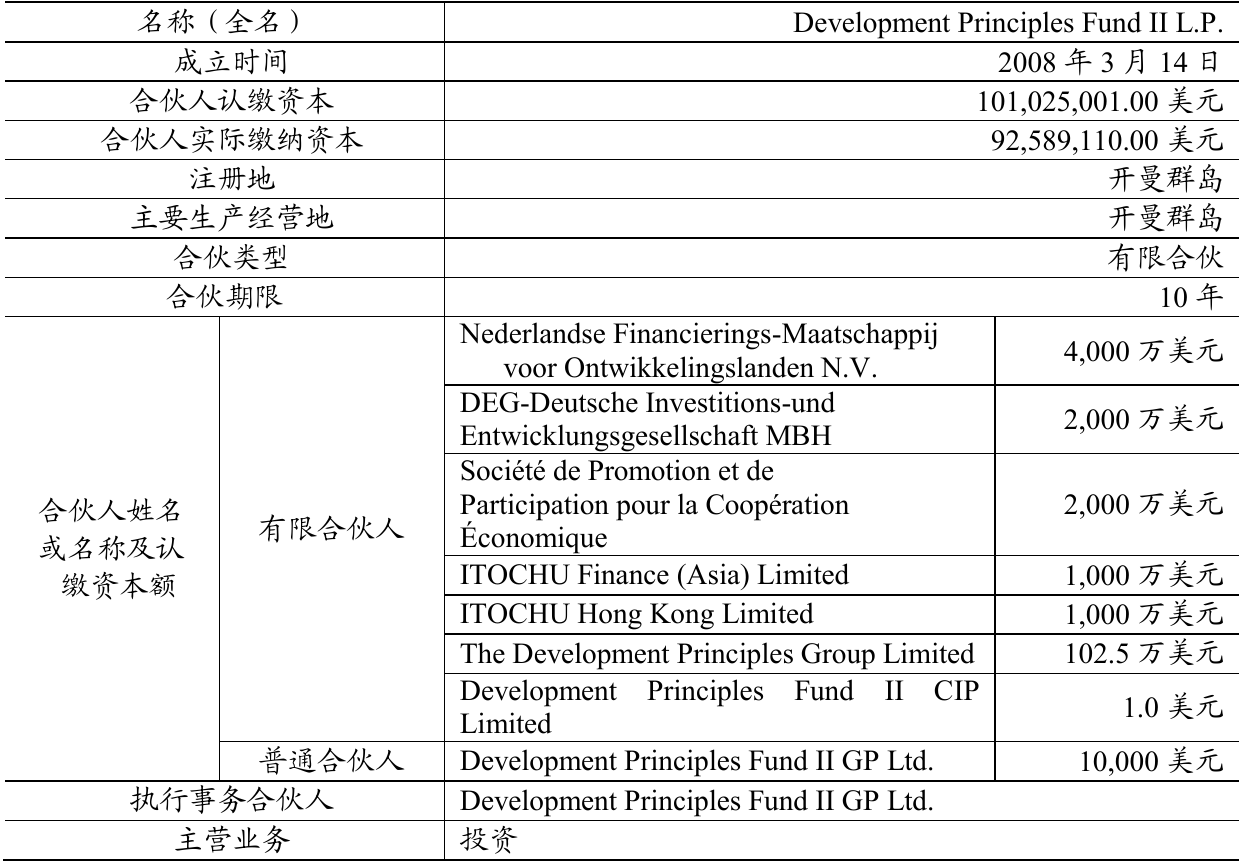}
\end{minipage}%
}%

\subfigure[In this profitability table, the first column is left-aligned for legibility and the other three columns are right-aligned because only then can the decimal points be aligned.]{
\begin{minipage}[t]{\linewidth}
\centering
\includegraphics[width=\linewidth]{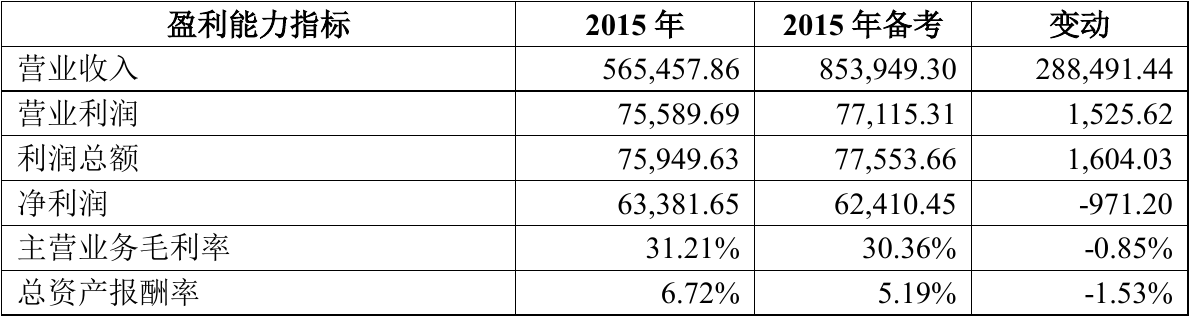}
\end{minipage}%
}%
\centering
\caption{Typical examples of unevenly distributed horizontal cells in financial tables.}
\label{fig::tbl}
\end{figure}

Meanwhile, we notice a higher accuracy in vertical prediction than in horizontal prediction on FinTab. It is possibly caused by the uneven distribution of cells within a row of financial tables. Figure \ref{fig::tbl} gives some typical examples. In Figure \ref{fig::tbl} (a), the nodes in the first 8 rows are distributed extremely far in horizontal direction. In Figure \ref{fig::tbl} (b), when calculating K nearest neighbors for the first column, many vertical relations will be included, but very few horizontal relations, especially when K is small. These situations are rather rare in academic tables but not uncommon in financial reports.

In Table \ref{tab::val-test}, we give the accuracy results of GFTE on different datasets, namely on the SciTSR test dataset for validation and on our FinTab dataset for test. It could be observed that our model reaches rather high accuracy on SciTSR validation dataset, which implies that our algorithm works well as a baseline given enough training data. In addition, although FinTab dataset have completely different sources and properties from SciTSR dataset, GFTE still achieves good results, which suggests that our model has certain robustness.

\begin{table}[htbp]
\caption{ACCURACY RESULTS OF BOTH VERTICAL AND HORIZONTAL RELATIONS ON VALIDATION DATASET AND TEST DATASET.}
\begin{center}
\begin{tabular}{|p{3cm}|p{1.5cm}<{\centering}|p{1.5cm}<{\centering}|}
\hline
\multicolumn{1}{|m{3cm}}{Dataset} & \multicolumn{1}{|m{1.5cm}}{Horizontal prediction} & \multicolumn{1}{|m{1.5cm}|}{Vertical\quad prediction} \\
\hline
\rowcolor{Gray}
SciTSR test dataset & 0.954048 & 0.922423 \\
FinTab dataset & 0.861019 & 0.903031 \\
\hline
\end{tabular}
\label{tab::val-test}
\end{center}
\end{table}

In conclusion, the performance of GFTE is overall good yet still unsatisfactory. Possible reasons can be:
\begin{itemize}
    \item Our model is trained on English dataset, yet tested on Chinese dataset. This may lead to deviation in text feature.
    \item Financial tables may have different characteristics from academic tables.
\end{itemize}

\section{Conclusion}
In this paper, we disclose a standard Chinese financial dataset from PDF files for table extraction benchmark test, which is diverse, sufficient and comprehensive. With this novel dataset, we hope more innovative and fine-designed algorithms of table extraction will emerge. Meanwhile, we propose a GCN-based algorithm GFTE as a baseline. We also discuss its performance and some possible difficulties by extracting tables from financial files in Chinese, which can be overcome in future works.

\end{document}